\documentclass[runningheads]{llncs}
\usepackage{graphicx}
\usepackage{multirow}
\usepackage{stackengine}
\usepackage[table]{xcolor}
\usepackage{array}
\usepackage{framed,multirow}
\usepackage{color}
\usepackage{xcolor}
\usepackage[colorlinks = true,
            linkcolor = green,
            urlcolor  = blue,
            citecolor = blue,
            anchorcolor = blue]{hyperref}
\usepackage{multirow}
\usepackage{eurosym}
\usepackage{amssymb}
\usepackage{amsmath,amssymb}
\usepackage{latexsym}
\usepackage{lipsum}

\usepackage{url}
\usepackage{xcolor}
\usepackage{float}
\newcommand{\squeezeup}{\vspace{-2.5mm}}
\newcommand{\concat}{%
  \mathbin{{+}\mspace{-08mu}{+}}%
}

\definecolor{newcolor}{rgb}{.8,.349,.1}
\begin{document}
\title{A Deep Learning Approach for Real-Time 3D Human Action Recognition from Skeletal Data}
%
%
\author{Huy Hieu Pham\inst{1,2}\orcidID{0000-0003-4851-2518} \and 
Houssam Salmane\inst{1}\orcidID{0000-0002-0919-7482} \and
Louahdi Khoudour\inst{1}\orcidID{000-0002-5947-4302} \and
Alain Crouzil\inst{2}\orcidID{0000-0001-7040-2978} \and \\
Pablo Zegers\inst{3}\orcidID{0000-0003-3697-2525} \and
Sergio A. Velastin\inst{4,5}\orcidID{0000-0001-6775-1737}}

\authorrunning{Hieu Pham et al.}
\titlerunning{Deep Learning for Real-Time Action Recognition from Skeletal Data}
%
\institute{Cerema, Equipe-projet STI, 1 Avenue du Colonel Roche, 31400, Toulouse, France; \\ \email{\{huy-hieu.pham,louahdi.khoudour,houssam.salmane\}@cerema.fr} \and
Universit\'e Toulouse III - Paul Sabatier, Institut de Recherche en Informatique de Toulouse, F-31062 Cedex 9, Toulouse, France; \email{alain.crouzil@irit.fr}\\ \and
Aparnix, La Gioconda 4355, Santiago, Chile; \email{pablozegers@gmail.com}  \and
Cortexica Vision Systems Ltd., London, UK \and 
Queen Mary University of London and Department of Computer Science, University Carlos III of Madrid, Madrid, Spain; \email{sergio.velastin@ieee.org}}
\maketitle              
\begin{abstract}
We present a new deep learning approach for real-time 3D human action recognition from skeletal data and apply it to develop a vision-based intelligent surveillance system. Given a skeleton sequence, we propose to encode skeleton poses and their motions into a single RGB image. An Adaptive Histogram Equalization (AHE) algorithm is then applied on the color images to enhance their local patterns and generate more discriminative features. For learning and classification tasks, we design Deep Neural Networks based on the Densely Connected Convolutional Architecture (DenseNet) to extract features from enhanced-color images and classify them into classes. Experimental results on two challenging datasets show that the proposed method reaches state-of-the-art accuracy, whilst requiring low computational time for training and inference. This paper also introduces CEMEST, a new RGB-D dataset depicting passenger behaviors in public transport. It consists of 203 untrimmed real-world surveillance videos of realistic normal and anomalous events. We achieve promising results on real conditions of this dataset with the support of data augmentation and transfer learning techniques. This enables the construction of real-world applications based on deep learning for enhancing monitoring and security in public transport.

\end{abstract}
\section{Introduction} \label{sec1}
\noindent Human Action Recognition or HAR for short, plays a crucial role in many computer vision applications such as intelligent surveillance, human-computer interaction or robotics. Although significant progress has been achieved, detecting accurately what humans do in unknown videos is still a challenging task due to numerous challenges, e.g. viewpoint changes, intra-class variation, or surrounding distractions \cite{poppe2010survey}. At present, depth sensor-based HAR is considered as one of the best available methods for overcoming the above obstacles. Cost-effective depth sensors are able to provide 3D structural information of the human body, which is suitable for HAR task. In particular, most of these devices have integrated the real-time skeleton estimation algorithms \cite{shotton2013real} that are robust to surrounding distractions as well as invariant to camera viewpoints. Therefore, exploiting skeletal data for HAR opens up opportunities for addressing the limitations of RGB and depth modalities. In the literature of skeleton-based action recognition, there are two main issues that need to be solved. The first challenge is how to transform the raw skeleton sequences into an effective representation, which is able to capture the spatio-temporal dynamics of human motions. The second is to model and recognize actions using the motion representation obtained from skeletons. Previous works on this topic can be divided into two main groups: HAR based on hand-crafted features and HAR using deep learning models \cite{pham2015video,pham2019architectures}. The first group of methods extracts hand-crafted local features from skeleton joints and uses probabilistic graphical models such as Hidden Markov Model (HMM) \cite{lv2006recognition}, Conditional Random Field (CRF) \cite{han2010discriminative}, and Fourier Temporal Pyramid (FTP) \cite{Vemulapalli2014HumanAR} to model and classify actions. For instance, since the first work on 3D HAR from depth data was introduced \cite{Li2010ActionRB}, many methods for skeleton-based action recognition have been proposed \cite{han2010discriminative,wang2012mining,xia2012view,luo2013group,wang2014mining,Vemulapalli2014HumanAR,wu2014leveraging,Wang2016GraphBS}. The common characteristic of these approaches is that, they extract the geometric features from the 3D coordinates of the skeleton joints and model their temporal information by a generative model. Although promising results have been achieved, most of these approaches are shallow, data-dependent and require a lot of feature engineering. E.g., they require pre-processing input data in which the skeleton sequences need to be segmented or aligned. In contrast, we propose a skeleton-based representation and a learning framework for 3D HAR that learns to recognize actions from the raw skeletons in an end-to-end manner, without dependence on the length of actions.\\[0.1cm]
\hspace*{0.5cm}The second group considers skeleton-based action recognition as a time-series problem and proposes to use Recurrent Neural Networks with Long Short-Term Memory units (RNN-LSTMs) \cite{hochreiter1997long} to analyze the temporal evolutions of skeletons. They are considered as the most popular deep learning based approach for the HAR task from skeletons and have achieved high-level performance \cite{Du2015HierarchicalRN,veeriah2015differential,zhu2016co,Shahroudy2016NTURA,Liu2016SpatioTemporalLW,liu2017global,si2018skeleton}. The temporal evolutions of skeletons are in fact spatio-temporal patterns. Thus, they can be modeled by memory cells in the structure of RNN-LSTMs. However, RNN-LSTM based methods tend to overemphasize the temporal information and lose spatial information of skeletons \cite{Du2015HierarchicalRN} -- an important characteristic for 3D HAR. Another limitation of RNN-LSTM networks is that they just model the overall temporal dynamics of actions without considering the detailed temporal dynamics of them \cite{lee2017ensemble}. Additionally, this approach considers skeletons as a kind of low-level feature by feeding raw skeletal data directly into the network. The huge number of input features makes RNN-LSTMs complex, time-consuming and easily lead to overfitting. Furthermore in many cases, RNN-LSTMs act as a classifier, which cannot extract high-level features for the HAR problem \cite{sainath2015convolutional}. In this paper, we propose a CNN-based method to extract rich geometric motion features from skeleton sequences and model various temporal dynamics, including both short-term and long-term actions.
\begin{figure*}[htb]
\begin{minipage}[b]{1.0\linewidth}
  \centering
  \centerline{\includegraphics[width=12.65cm,height=2.5cm]{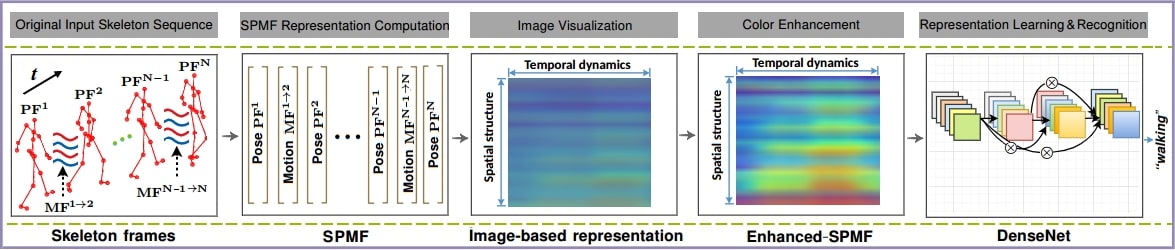}}
\end{minipage}
\caption{Overview of the proposed approach for real-time 3D action recognition from skeletal data. Each skeleton sequence is encoded as a single RGB image via a skeleton-based representation called SPMF \cite{Hieu_et_al_2018}. The SPMF is built from \textit{Pose Feature} vectors (PFs) and \textit{Motion Feature} vectors (MFs), which are estimated from the 3D coordinates of skeletons. A color enhancement technique \cite{pizer1987adaptive} is adopted to enhance the local textures of SPMF to form the enhanced motion maps, namely Enhanced-SPMF. Finally, they are fed to a deep network for learning image features and performing action classification.}
\label{fig:1}
\end{figure*}
In contrast to the existing approaches, we aim to build an end-to-end deep learning framework for real-time action recognition from skeleton sequences. We believe that an effective motion representation is the key factor influencing recognition performance. Therefore, we propose to encode human poses and motions extracted from the 3D coordinates of skeleton joints into color images. These color-coded images are then enhanced in their local textures by an Adaptive Histogram Equalization (AHE) algorithm \cite{pizer1987adaptive} before feeding into Deep Convolutional Neural Networks (D-CNNs), which are built based on the DenseNet architecture \cite{Huang2016DenselyCC}. Before that, a smoothing filter is applied to reduce the effects of noise on the input skeletal data. The overview of the proposed method is illustrated in Fig.~\ref{fig:1}. Generally speaking, four hypotheses that motivate us to build a skeleton-based representation and design DenseNets for 3D HAR include: (\textbf{1}) human actions can be correctly represented via movements of the skeleton \cite{johansson1973visual}; (\textbf{2}) spatio-temporal evolutions of skeletons can be transformed into color images -- a kind of 3D tensor that can be effectively learned by D-CNNs \cite{bilen2018action,ding2017investigation,choutas2018potion}. This hypothesis was proved in our previous studies \cite{pham2017learning,Hieu_et_al_2018,Pham2018learning,pham2018exploiting,pham2019spatio,pham2020unified}; (\textbf{3}) compared to RGB and depth modalities, skeletal data has high-level information with much less complexity. This makes the learning model much simpler and requiring less computation, allowing us to build real-time deep learning framework for HAR task; (\textbf{4}) DenseNet is currently one of the most effective CNN architecture for image recognition. It has a densely connected structure allowing maximal information flow and facilitates features reuse as each layer in its architecture has direct access to the features from previous layers. This helps DenseNet to improve its learning performance. Therefore, we explore and optimise this architecture for learning and recognizing human actions on the proposed image-based representation.

The main contributions of this work are three-fold. \textbf{First}, we introduce Enhanced-SPMF (Enhanced Skeleton Pose-Motion Feature) -- a 3D motion representation for HAR tasks (Sec.~\ref{enhanced-SPMF}). This is an extended representation of SPMF, which was presented in our previous work \cite{Hieu_et_al_2018}. The new representation aims to improve the efficiency of the SPMF by using a smoothing filter on input skeleton sequences and a color enhancement technique that could make the proposed Enhanced-SPMF more robust and discriminative. An ablation study on the Enhanced-SPMF demonstrates that the new representation leads to better overall action recognition performance than the SPMF. \textbf{Second}, we introduce an end-to-end deep framework based on D-CNNs\footnote{Codes and models are available on our GitHub project at \url{https://bit.ly/2EC9vj9}.} for learning and recognizing actions from the Enhanced-SPMFs (Sec.~\ref{DenseNet}). This approach is general in the sense that it can be applied to other data modalities, e.g. mocap data or the output of 3D pose estimation algorithms. The proposed method is evaluated on two highly competitive benchmark datasets and achieved state-of-the-art performance on both these two benchmark tasks with high computational efficiency (Sec.~\ref{sect:Experiments}). \textbf{Finally}, we collect and introduce a new RGB-D dataset consisting of real-world surveillance videos for analyzing anomalous and normal events in public transport. Experimental results show that the proposed method achieves promising performance in realistic conditions (Sec.~\ref{sect:Experiments}). 

The rest of this paper is organized as follows: Sec.~\ref{enhanced-SPMF} presents the proposed skeleton-based representation. The proposed deep learning framework is presented in Sec.~\ref{DenseNet}. Datasets and experiments are provided in Sec.~\ref{sect:Experiments}, including a description of our dataset and the obtained results. Sec.~\ref{sect:Conclusions} concludes the paper. \\[-0.7cm]
\section{Enhanced Skeleton Pose-Motion Feature \\[-0.6cm]} \label{enhanced-SPMF}
\begin{figure}
\centering
\includegraphics[width=10cm,height=2cm]{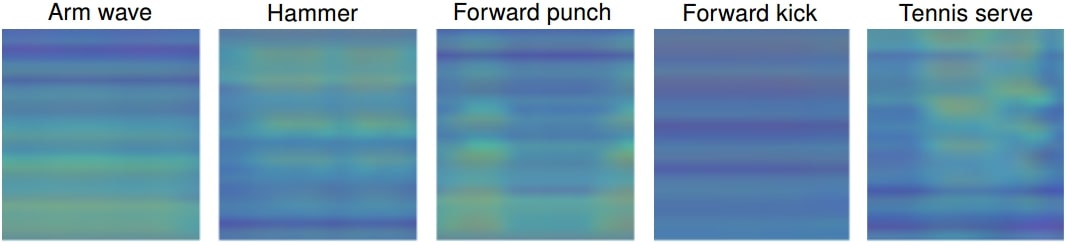}
\caption{Some SPMF representations obtained from the MSR Action3D dataset \cite{Li2010ActionRB}. The change in color reflects the change of distance and orientation between the joints.}
\label{fig:SPMF}
\end{figure}

\noindent One of the major challenges in exploiting D-CNNs for skeleton-based action recognition is how a skeleton sequence could be effectively represented and fed to the deep networks. As D-CNNs work well on still images \cite{lecun2015deep}, our idea therefore is to encode the spatial and temporal dynamics of skeletons into 2D images \cite{Pham2018learning,pham2018exploiting}. Two essential elements for describing an action are static poses and their temporal dynamics. As shown by Zhang et \textit{al.} in \cite{Zhang2017OnGF}, the combination of too many geometric features will lead to lower performance than using only a single feature or several main features. Moreover, joint features such as joint distance and joint motion are stronger features than others \cite{yun2012two}. Hence, we decide to transform these two important elements into the static spatial structure of a color image. The details of this idea are explained in our previous work \cite{Hieu_et_al_2018}, in which the spatio-temporal patterns of a skeleton sequence can be encoded into a single color image as a global representation, namely SPMF, via pose and motion feature vectors. Due to the limited space available, detailed description of the SPMF is not included. Instead, we refer the interested readers to \cite{Hieu_et_al_2018} for further technical details. Fig. \ref{fig:SPMF} shows some SPMF representations in form of an image-based representation obtained from MSR Action3D dataset \cite{Li2010ActionRB}. \\[-0.65cm]

\begin{figure}
\centering
\includegraphics[width=10cm,height=2cm]{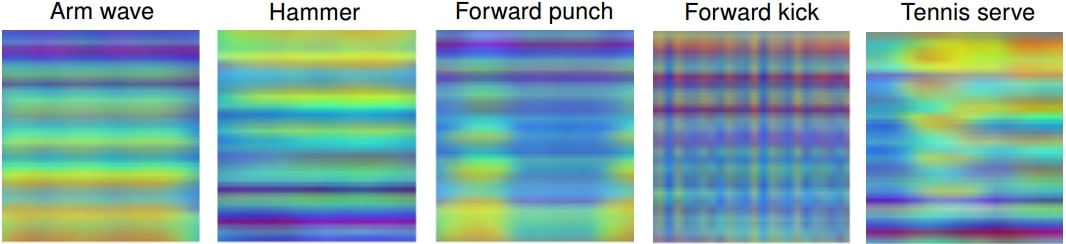}
\caption{The corresponding Enhanced-SPMF representations after applying the AHE algorithm \cite{pizer1987adaptive}. This color enhancement step could make the proposed Enhanced-SPMF more robust and discriminative for the representation learning phase with D-CNNs later. }
\label{fig:Enhanced-SPMF}
\end{figure}

The color images obtained after the process of encoding mainly reflect the spatio-temporal distribution of skeleton joints. We observe that these images are represented by close contrast values, as can be seen in as Fig. \ref{fig:SPMF}. In this case, a color enhancement method can be useful for increasing the contrast of these representations and highlighting the texture and edges of motion maps. This helps to better distinguish similar actions. Therefore, it is necessary to enhance the local features on the generated color images. The Adaptive Histogram Equalization (AHE) \cite{pizer1987adaptive} is a common approach for this task. This technique is capable of enhancing the local features of an image. Mathematically, let $\textbf{\textit{I}}$ be a given image, represented as a $r$-by-$c$ matrix of integer pixels with intensity levels in the range $[0,L-1]$. The histogram of image $\textbf{\textit{I}}$ will be defined by $H_k = \textbf{n}_k$, where $\textbf{n}_k$ is the number of pixels with intensity $k$ in $\textbf{\textit{I}}$. Hence, the probability of occurrence of intensity level $k$ in $\textbf{\textit{I}}$ is 
\begin{equation}
p_k =  \dfrac{\textbf{n}_k}{r \times c}, \hspace{0.2cm} (k = 0, 1, 2, ..., L-1). 
\end{equation}
The histogram equalized image will be formed by transforming the
pixel intensities, $n$, of  $\textbf{\textit{I}}$ by the function \squeezeup
\begin{equation}
T(n) =\texttt{floor} ((L-1)\sum_{k=0}^{n}p_k), \hspace{0.2cm} (n = 0, 1, 2, ..., L-1),
\end{equation}
The Histogram Equalization (HE) method is used for increasing the global contrast of the image. However, it cannot solve the problem of increasing the local contrast. To do this, the image needs to be divided into $\mathcal{R}$ regions and the HE is then applied in each region. This technique is called the Adaptive Histogram Equalization (AHE). Fig.~\ref{fig:Enhanced-SPMF} shows samples of the enhanced motion maps with $\mathcal{R} = 8$, which we refer to it as Enhanced-SPMF for some actions from the MSR Action 3D dataset \cite{Li2010ActionRB}.\\[-0.65cm]
\section{Deep learning model}\label{DenseNet} 
\noindent This section reviews the key ideas behind the DenseNet architecture and presents the proposed deep networks for recognizing actions from the Enhanced-SPMFs.\\[-0.65cm]
\subsection{DenseNet review}
\noindent DenseNet \cite{Huang2016DenselyCC}, a recently proposed CNN model, has some interesting properties. Each layer is connected to all the others within a dense block and all layers can access feature maps from their preceding layers. Besides, each layer receives direct information flow from the loss function through shortcut connections. These properties make DenseNet less prone to overfitting for supervised learning problems. Traditional CNN architectures use the output feature maps $\textbf{x}_{l-1}$ of the $({l-1})^\text{th}$ layer as input to the ${l}^\text{th}$ layer and learn a mapping function $\textbf{x}_{l} = \mathcal{H}_l({\textbf{x}_{l-1})}$. Here, $\mathcal{H}_l(\cdot)$ is a non-linear transformation that is usually implemented by a series of  operations such as Convolution (\textbf{Conv}.), Rectified Linear Unit (\textbf{ReLU}) \cite{glorot2011deep}, Pooling, and Batch Normalization (\textbf{BN}) \cite{ioffe2015batch}. When increasing the depth of the network, the problem of optimization becomes complex due to the vanishing-gradient problem and the degradation phenomenon \cite{he2015convolutional}. To solve these problems, \cite{He2016DeepRL} introduced ResNet. The key idea behind the ResNet architecture is to add shortcut connections that bypass the non-linear transformations $\mathcal{H}_l(\cdot)$ with an identity function $\textit{id}(\textbf{x}) = \textbf{x}$. Inspired by the philosophy of ResNet, to maximize information flow through layers, Huang et \textit{al}. \cite{Huang2016DenselyCC} proposed DenseNet in which the $l^\text{th}$ layer in a dense block receives the feature maps of all preceding layers as inputs. That means
\begin{equation}
\textbf{x}_l =  \mathcal{H}_l( \texttt{concat}[\textbf{x}_0, \textbf{x}_1, \textbf{x}_2, ... , \textbf{x}_{l-1}]),
\end{equation}
where $\texttt{concat}[\textbf{x}_0, \textbf{x}_1, \textbf{x}_2, ... , \textbf{x}_{l-1}]$ is a single tensor constructed by concatenation of the previous layer output feature maps. All layers receive direct supervision signal from the loss function through the shortcut connections. Therefore, DenseNets are easy to optimize and resistant to overfitting. In DenseNet, multiple dense blocks are connected via transition layers. Each block with its transition layer produces $\textit{k}$ feature maps and the parameter $\textit{k}$ is called as the \textit{``growth rate''} of the network. The function $\mathcal{H}_l(\cdot)$ in the original work \cite{Huang2016DenselyCC} is a composite function of three consecutive layers: \textbf{BN}-\textbf{ReLU}-\textbf{Conv}. 
\begin{figure}
\centering
\includegraphics[width=12.8cm,height=3.5cm]{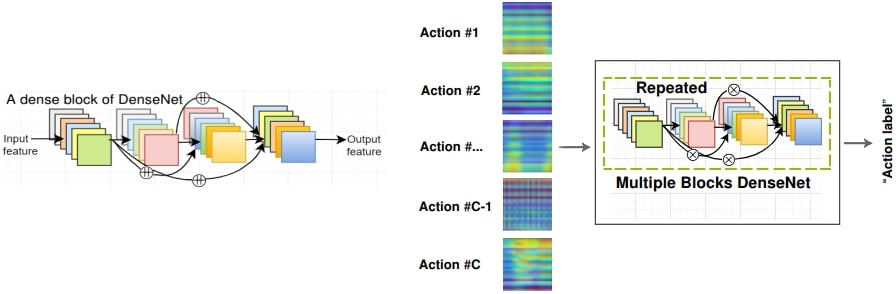}
\squeezeup
\caption{A DenseNet block (\textbf{left}). The symbols {\scriptsize \textcircled{$\concat$}} denotes the concatenation operator. We explore and optimize this architecture for learning and recognizing human actions on the proposed image-based representation (\textbf{right}).}
\label{fig:}
\end{figure}
\subsection{Network design}
\noindent We design D-CNNs based on the DenseNet architecture \cite{Huang2016DenselyCC} to learn and classify actions on the Enhanced-SPMF. To study how performance varies with architecture size, we test three different configurations of DenseNet: $\{$DenseNet-16, \textit{k} = 12\}$ ; \{\text{DenseNet-28}, \textit{k} = 12\}$; and $\{\text{DenseNet-40}, \textit{k} = 12\}$. Here, the numbers 16, 28, 40 refer to the depth of the network and \textit{k} is the network growth rate. For computational efficiency, we use three dense blocks on $32\times32$ input images. The $\mathcal{H}_l(\cdot)$ function is implemented by a sequence of layers: Batch Normalization (\textbf{BN}), advanced activation layer named Exponential Linear Units (\textbf{ELU}) \cite{clevert2015fast} and $3\times3$ Convolution (\textbf{Conv}). Dropout \cite{clevert2015fast} with a rate of 0.2 is used after each \textbf{Conv}. to prevent overfitting. The proposed networks can be trained in an end-to-end manner by gradient descent using Adam update rule \cite{kingma2014adam}. During training, we minimize a cross-entropy loss function between the true action label $\textbf{y}$ and the predicted action $\hat{\textbf{y}}$ over the training samples $\mathcal{X}$, by solving the following optimization problem \squeezeup
\begin{equation}
\stackunder{Arg min}{$\mathcal{W}$} (\mathcal{L}_{\mathcal{X}}(\textbf{y},\hat{\textbf{y}})) = \stackunder{Arg min}{$\mathcal{W}$} \left (   - \frac{1}{M} \sum_{i = 1}^{M}   \sum_{j = 1}^{C} \textbf{y}_{ij} \log \hat{\textbf{y}}_{ij}\right ),
\end{equation}
where $\mathcal{W}$ is the set of weights that will be learned by the model, $M$ denotes the number of samples in training set $\mathcal{X}$ and $C$ is the number of action classes. 
\squeezeup
\section{Experiments} \label{sect:Experiments}
\squeezeup
\noindent The proposed method is first evaluated on two challenging datasets: the MSR Action3D and NTU RGB+D (Sec.~\ref{datasets-and-settings}). We then introduce the CEMEST dataset\footnote{Created by Cerema and Tiss\'eo public transport in France and available for research purposes from \url{https://bit.ly/2SNbrdE}.} and report experimental results on this dataset. The implementation details of the proposed D-CNNs are also provided in this section (Sec.~\ref{implementation-details}).
\squeezeup
\subsection{Datasets and settings \label{datasets-and-settings}}

\noindent \textbf{MSR Action3D dataset} \cite{Li2010ActionRB}: This dataset contains 20 actions performed by 10 subjects. Each skeleton is composed of 20 key joints. Experiments were conducted on 557 action sequences. We follow the protocol proposed by \cite{Li2010ActionRB}. Specifically, the whole dataset is divided into three subsets: AS1, AS2 and AS3. For each subset, five subjects are selected for training and the rest are used for testing (see Supplemental Material). Data augmentation techniques including random cropping, vertically flipping, and Gaussian filtering have been applied on this dataset.\\[0.2cm]
\noindent \textbf{NTU RGB+D dataset} \cite{Shahroudy2016NTURA}: This Kinect 2 captured dataset is a very large-scale RGB+D dataset. It is currently the largest and state-of-the-art dataset that provides skeletal data for 3D HAR. The NTU RGB+D has more than 56 thousand video samples, 4 millions frames, collected from 40 distinct subjects for 60 different action classes. Each skeleton contains the 3D coordinates of 25 body joints. The authors of this dataset suggested two different evaluation criteria, including Cross-Subject and Cross-View evaluations. For the Cross-Subject setting, the sequences performed by 20 subjects are used for training and the rest sequences are used for testing. In Cross-View setting, the sequences provided by cameras 2 and 3 are used for training while sequences from camera 1 are used for testing (see Supplemental Material). Due to the very large-scale nature of the NTU RGB+D dataset, we do not apply any data augmentation technique on this dataset.
\begin{figure}
\centering
\includegraphics[width=4.15cm,height=2.5cm]{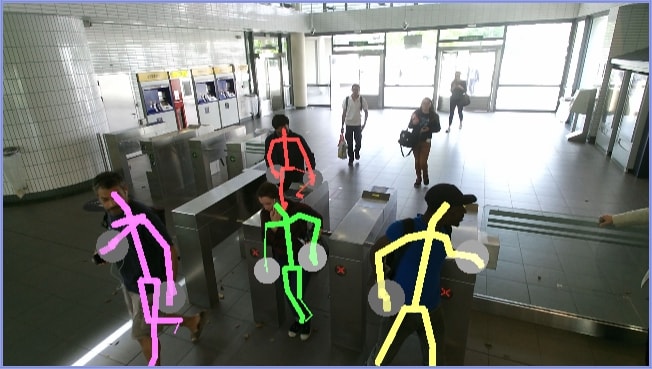}
\includegraphics[width=4.15cm,height=2.5cm]{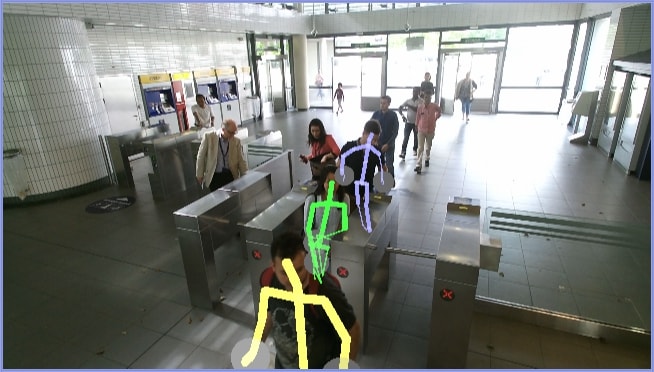}
\includegraphics[width=4.15cm,height=2.5cm]{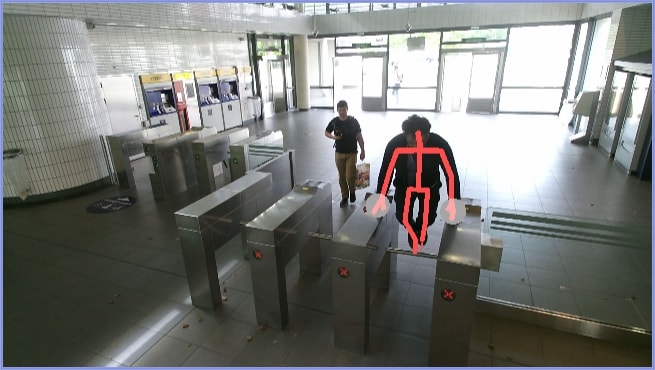}\\
{\footnotesize(a) \hspace{3cm} (b) \hspace{3cm}  (c)} \\
\includegraphics[width=4.15cm,height=2.5cm]{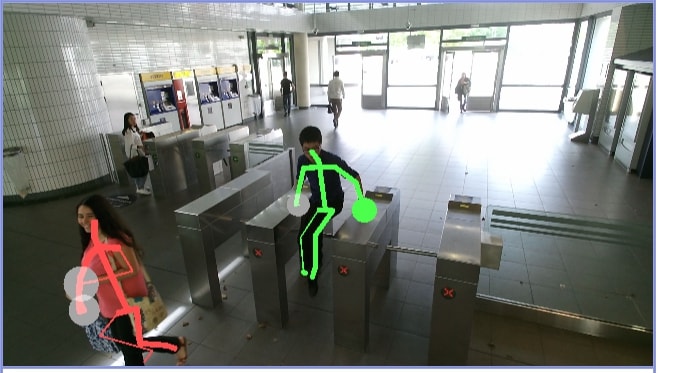}
\includegraphics[width=4.15cm,height=2.5cm]{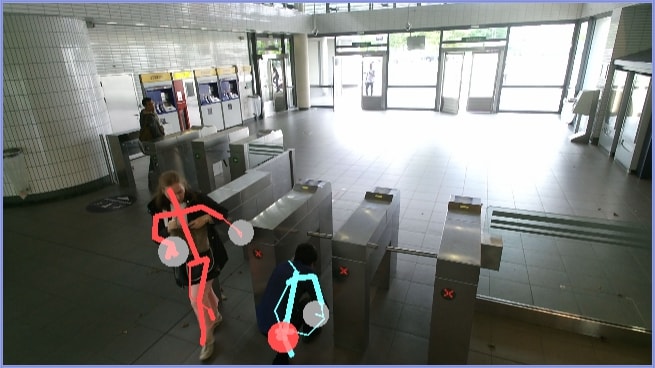}
\includegraphics[width=4.15cm,height=2.5cm]{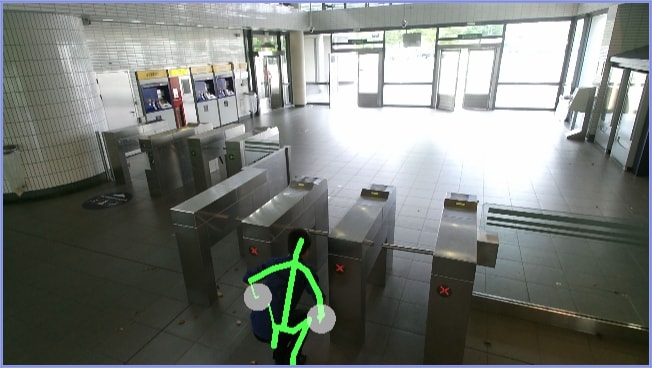}\\
{\footnotesize(d) \hspace{3cm} (e) \hspace{3cm} (f)}\\
\squeezeup
\caption{Some samples from the CEMEST dataset: (a), (b) \textit{crossing over the barriers}, (c), (d) \textit{jumping over the ticket barriers} and (e), (f) \textit{sneaking under ticket barriers}.}
\label{fig:tisseo-cerama}
\end{figure}\\[0.2cm]
\textbf{CEMEST dataset}: We have collected a new RGB-D dataset, called CEMEST (CErema MEtro STation dataset) using Kinect v2 sensor and carried out experiments on this dataset to verify the effectiveness of the proposed method on a real-world dataset. The CEMEST was made at a metro station in France without any control of the passenger behavior as well as illumination. It contains three actions including both normal and abnormal behaviors: \textit{crossing normally over the barriers}, \textit{jumping over the barriers}, and \textit{sneaking under the barriers}.  These three behaviors are taken into account for acquisition because they have a significant impact on monitoring and management in public transport. As an example, the French National Railway Company (SNCF) reported that they lost \euro 500 million every year through people trying to cheat the ticket system \cite{sncf}. In summary, this dataset provides RGB, depth and skeletal data. The skeleton sequences are extracted by Kinect SDK with 25 key joints for each subject, at a frame rate of 30 FPS.  All recorded sequences are manually segmented and labeled. Fig.~\ref{fig:tisseo-cerama} shows some samples from the CEMEST. We carried out two experimental evaluations on this dataset. In the first setting, we randomly chose 67\% of the data as training set and the remaining 33\% is used for testing. In the second setting, the proposed networks are trained on a combination dataset, which is created from a portion of the MSR Action3D \cite{Li2010ActionRB} and NTU RGB+D \cite{Shahroudy2016NTURA} datasets (see Supplementary Material for more details). To ensure the number of samples in each action class is balanced, we augmented samples in the MSR Action3D to match the size of the larger dataset. The pre-trained model is then deployed on the CEMEST dataset in the hope that transfer learning will help to solve overfitting problem when training on small dataset. In both experiments, data augmentation (i.e. cropping, flipping, Gaussian filtering) has been used. \\[-0.65cm]
\subsection{Implementation details} 
\label{implementation-details}
\noindent The Enhanced-SPMFs are computed directly from skeletons without using a fixed number of frames. The proposed DenseNets were implemented in Python using Keras. For training, we use mini-batches of 64 images. The weights are initialized as \cite{He2015DelvingDI}. Adam optimizer \cite{kingma2014adam} is used with an initial learning rate $\eta$  = 3e-4. All networks are trained for 250 epochs from scratch.
\squeezeup
\subsection{Experimental results and evaluation}
\label{experimental-results}
\begin{figure}[H]
\centering
 \centerline{\includegraphics[width=5cm,height=3cm]{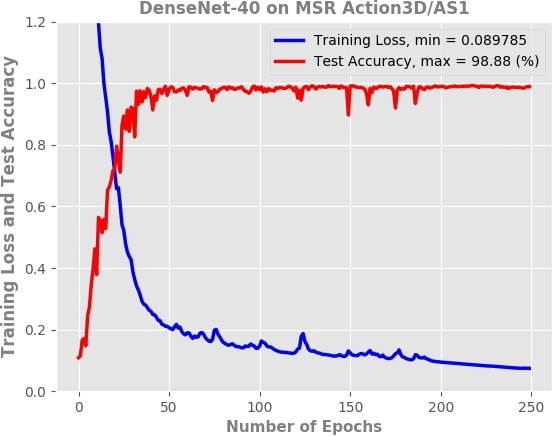}\hspace{1cm}\includegraphics[width=5cm,height=3cm]{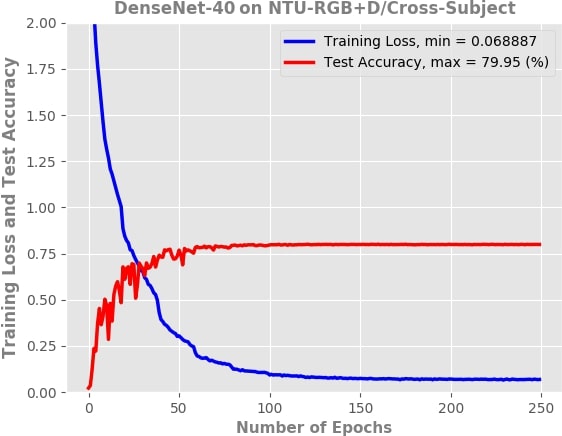}}
\caption{\label{fig:5} Learning curves of the proposed DenseNet-40 network on the Enhanced-SPMFs obtained from the MSR Action3D and NTU RGB+D.}
\label{fig:accuracy}
\end{figure}

\begin{figure*}
\begin{minipage}[b]{1.0\linewidth}
  \centering
  \centerline{\includegraphics[width=4.5cm,height=3.5cm]{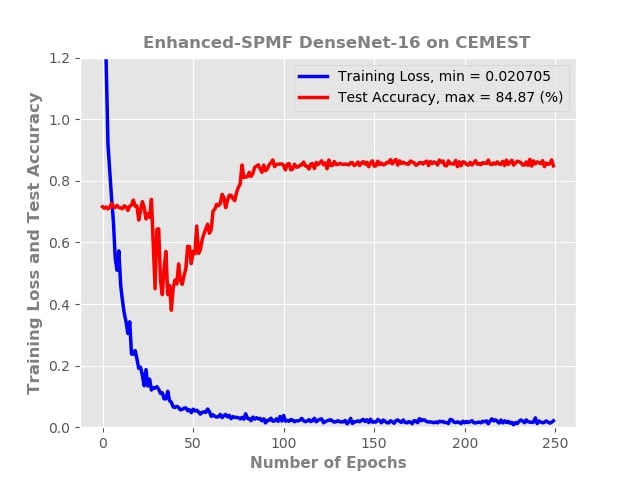}\includegraphics[width=4.5cm,height=3.5cm]{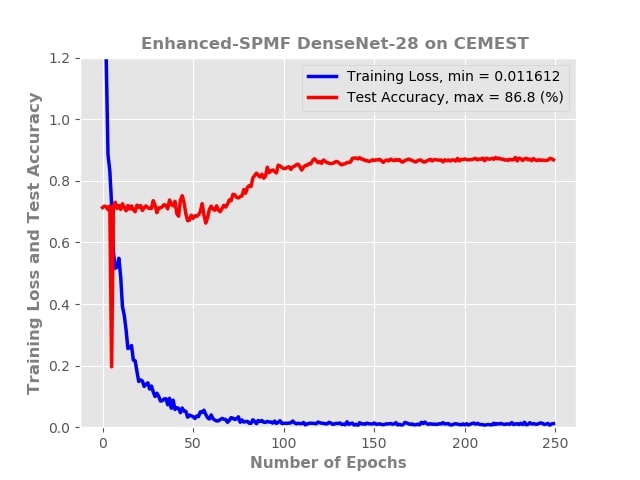}\includegraphics[width=4.5cm,height=3.5cm]{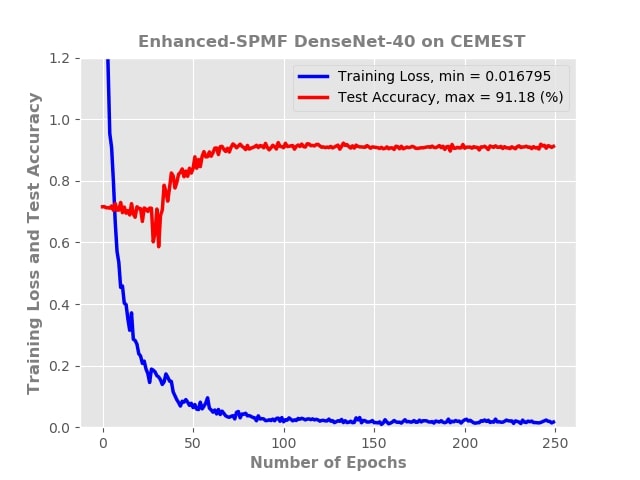}}
\hspace*{0.3cm}  (a) \hspace{3.8cm}   (b) \hspace{4cm}   (c)
  \centerline{ \includegraphics[width=4.5cm,height=3.5cm]{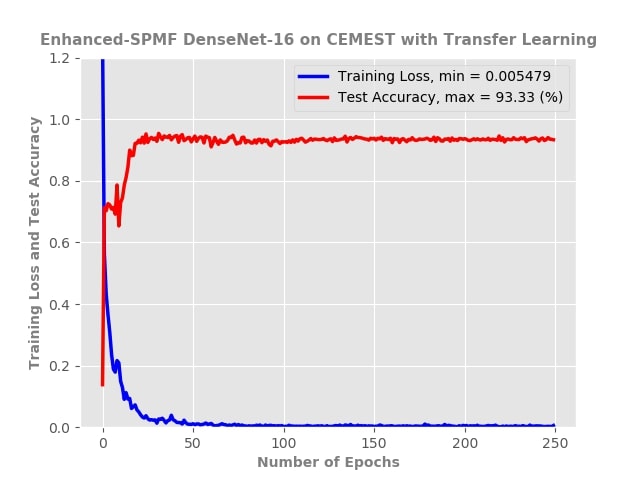}\includegraphics[width=4.5cm,height=3.5cm]{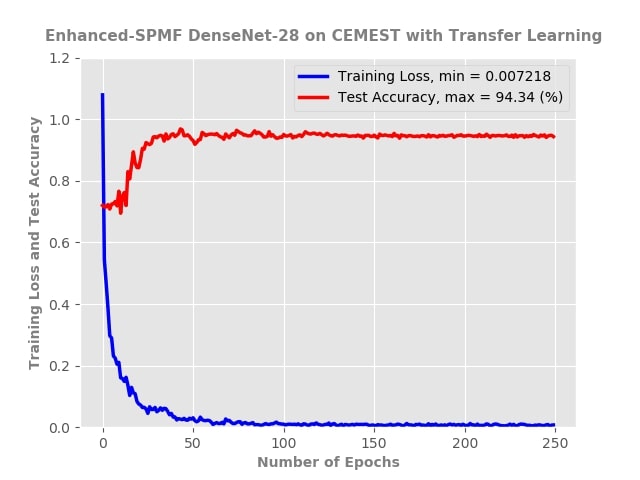}\includegraphics[width=4.5cm,height=3.5cm]{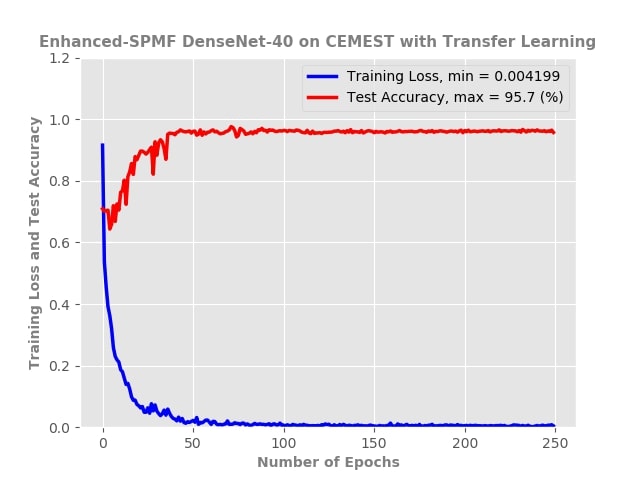}}
\hspace*{0.3cm}  (d) \hspace{3.8cm}  (e)  \hspace{4cm}  (f)
\end{minipage}
\caption{Learning curves of the three proposed deep networks (DenseNet-16, DenseNet-28, DenseNet-40) on CEMEST dataset when trained from scratch (a)-(b)-(c) and fine-tuned on CEMEST dataset (d)-(e)-(f). Our best configuration (DenseNet-40) achieved an accuracy of 91.18\% when trained on the CEMEST dataset from scratch. With the support of transfer learning, the proposed method reached an accuracy of 95.70\%, increasing the recognition accuracy by nearly 5\%.}
\label{fig:cemest}
\end{figure*}

\noindent Experimental results and comparison of the proposed method with existing approaches on the MSR Action3D dataset are summarized in Table~\ref{MSR-Action3D-accuracy}. The DenseNet-40 achieves an average accuracy of 99.10\% over three subsets, which outperformed previous approaches by \cite{Vemulapalli2014HumanAR,Wang2016GraphBS,Du2015HierarchicalRN,Li2010ActionRB,Chen2013RealtimeHA,WengSpatioTemporalNN,Xu2015SpatioTemporalPM} and surpassed our previous work on SPMF \cite{Hieu_et_al_2018}. Fig.~\ref{fig:5} (\textit{left}) shows an example of the learning curves of the network during training on this dataset. \\[-0.65cm]
\begin{table}
\centering
\caption{\label{MSR-Action3D-accuracy} Experimental results and comparison with the state-the-art approaches on the MSR Action3D dataset \cite{Li2010ActionRB}. The best accuracies are in \textcolor{blue}{\textbf{bold-blue}}. Results that surpass previous works are in \textbf{bold}.}
\begin{tabular}{p{5cm}p{1.8cm}p{1.8cm}p{1.8cm}p{1.8cm}}
\hline
{\footnotesize \textbf{Method} (protocol of \cite{Li2010ActionRB})}  & {\footnotesize  \textbf{AS1}} & {\footnotesize \textbf{AS2}} & {\footnotesize   \textbf{AS3}} &  {\footnotesize \textbf{Aver.}}\\
\hline
{\scriptsize Bag of 3D Points \cite{Li2010ActionRB}}  & {\scriptsize 72.90\%} & {\scriptsize 71.90\%} & {\scriptsize 71.90\%}  & {\scriptsize 74.70\%}\\
{\scriptsize \cellcolor{gray!50} Depth Motion Maps \cite{Chen2013RealtimeHA}}  & {\scriptsize \cellcolor{gray!50} 96.20\%} & {\scriptsize \cellcolor{gray!50} 83.20\%}  & {\scriptsize \cellcolor{gray!50} 92.00\%}  &  {\scriptsize \cellcolor{gray!50} 90.47\%} \\
{\scriptsize  Bi-LSTM \cite{tanfous2018coding} } & {\scriptsize 92.72\%} &  {\scriptsize 84.93\%}  &  {\scriptsize 97.89\%} &   {\scriptsize 91.84\%}\\
{\scriptsize \cellcolor{gray!50} Lie Group \cite{Vemulapalli2014HumanAR}} & \cellcolor{gray!50} {\scriptsize 95.29\%} & \cellcolor{gray!50} {\scriptsize 83.87\%}  & \cellcolor{gray!50} {\scriptsize 98.22\%}  & \cellcolor{gray!50} {\scriptsize 92.46\%} \\
{\scriptsize  Hierarchical RNN \cite{Du2015HierarchicalRN}} & {\scriptsize   99.33\%} & {\scriptsize   94.64\%}  & {\scriptsize   95.50\%}  & {\scriptsize 94.49\%} \\
{\scriptsize \cellcolor{gray!50}  Graph-Based Motion \cite{Wang2016GraphBS}}  & {\scriptsize \cellcolor{gray!50}   93.60\%} & {\scriptsize  \cellcolor{gray!50}  95.50\%}  & {\scriptsize  \cellcolor{gray!50}  95.10\%}  &  \cellcolor{gray!50} {\scriptsize 94.80\%}\\
{\scriptsize ST-NBNN \cite{WengSpatioTemporalNN}} & {\scriptsize 91.50\%} & {\scriptsize 95.60\%} & {\scriptsize 97.30\%} &  {\scriptsize 94.80\%}\\
{\scriptsize \cellcolor{gray!50} ST-NBMIM \cite{weng2018discriminative}} & {\scriptsize \cellcolor{gray!50} 92.50\%}  & {\scriptsize \cellcolor{gray!50} 95.60\%} &  {\scriptsize \cellcolor{gray!50} 98.20\%} &  {\scriptsize \cellcolor{gray!50} 95.30\%}\\
{\scriptsize  S-T Pyramid \cite{Xu2015SpatioTemporalPM}} & {\scriptsize  \textcolor{blue}{\textbf{99.10\%}}} & {\scriptsize  92.90\%}  & {\scriptsize   96.40\%}  &   {\scriptsize 96.10\%} \\
{\scriptsize \cellcolor{gray!50} SPMF \cite{Hieu_et_al_2018}} & \cellcolor{gray!50} {\scriptsize  97.54\%}  & \cellcolor{gray!50} {\scriptsize 98.73\%} & \cellcolor{gray!50} {\scriptsize 99.41\%} & \cellcolor{gray!50} {\scriptsize 98.56\%}\\
\hline
{\scriptsize  Enhanced-SPMF DenseNet-16 (\textbf{ours})} & {\scriptsize  98.05\%} & {\scriptsize  98.38\%}  & {\scriptsize 98.80\%}   & {\scriptsize 98.41\%} \\
{\scriptsize \cellcolor{gray!50} Enhanced-SPMF DenseNet-28 (\textbf{ours})} & \cellcolor{gray!50} {\scriptsize 98.44\%} & \cellcolor{gray!50} {\scriptsize 98.47\%}  & \cellcolor{gray!50} {\scriptsize 99.18\%}  &  \cellcolor{gray!50} {\scriptsize \textbf{98.70}\%}\\
{\scriptsize Enhanced-SPMF DenseNet-40 (\textbf{ours})} & {\scriptsize  98.88\%} & {\scriptsize  \textcolor{blue}{\textbf{99.05\%}}}  &  {\scriptsize  99.24\%}   & {\scriptsize \textcolor{blue}{\textbf{99.10}\%}} \\ 
\hline \\[-0.65cm]
\end{tabular}
\end{table}
\begin{table}
\centering
\caption{\label{NTU-Accuracy} Experimental results and comparison with the state-the-art approaches on the NTU RGB+D dataset \cite{Shahroudy2016NTURA}. The best accuracies are in \textcolor{blue}{\textbf{bold-blue}}. Results that surpass previous works are in \textbf{bold}.}
\begin{tabular}{p{6cm}p{3.18cm}p{3.18cm}}
\hline
{\scriptsize \textbf{Method}  (protocol of \cite{Shahroudy2016NTURA})}  &  {\scriptsize \textbf{Cross-Subject}} &   {\scriptsize \textbf{Cross-View}}  \\
\hline
{\scriptsize Lie Group  \cite{Vemulapalli2014HumanAR}}  &   {\scriptsize  50.10\%} &   {\scriptsize  52.80\%}\\
{\scriptsize \cellcolor{gray!50} Hierarchical RNN \cite{Du2015HierarchicalRN}}  &   {\scriptsize \cellcolor{gray!50} 59.07\%} &  {\scriptsize \cellcolor{gray!50} 63.97\%}\\
{\scriptsize Dynamic Skeletons \cite{Hu2015JointlyLH}}  &  {\scriptsize 60.20\%} &  {\scriptsize 65.20\%}\\
{\scriptsize \cellcolor{gray!50} Two-Layer P-LSTM \cite{Shahroudy2016NTURA}}  &  {\scriptsize \cellcolor{gray!50} 62.93\%} &  {\scriptsize \cellcolor{gray!50} 70.27\%}\\
{\scriptsize ST-LSTM Trust Gates \cite{Liu2016SpatioTemporalLW}}  &  {\scriptsize 69.20\%} &  {\scriptsize 77.70\%}\\
{\scriptsize \cellcolor{gray!50} Geometric Features \cite{Zhang2017OnGF}}  &  {\scriptsize \cellcolor{gray!50} 70.26\%} &  {\scriptsize \cellcolor{gray!50} 82.39\%}\\
{\scriptsize Two-Stream RNN \cite{Wang2017ModelingTD}}  &   {\scriptsize 71.30\%} &  {\scriptsize 79.50\%}\\
{\scriptsize \cellcolor{gray!50} Enhanced Skeleton \cite{Liu2017EnhancedSV}} &   \cellcolor{gray!50} {\scriptsize 75.97\%} &  \cellcolor{gray!50}  {\scriptsize 82.56\%}\\
{\scriptsize GCA-LSTM \cite{liu2018skeleton}}  &   {\scriptsize 76.10\%} &  {\scriptsize 84.00\%}\\
{\scriptsize \cellcolor{gray!50} SPMF \cite{Hieu_et_al_2018}}  &  \cellcolor{gray!50}  {\scriptsize 78.89\%} &  \cellcolor{gray!50} {\scriptsize 86.15\%}\\
\hline
{\scriptsize  Enhanced-SPMF DenseNet-16 (\textbf{ours})}  &  {\scriptsize 77.89\%} &  {\scriptsize \textbf{86.55\%}}\\
{\scriptsize  \cellcolor{gray!50} Enhanced-SPMF DenseNet-28 (\textbf{ours})}  &  \cellcolor{gray!50}  {\scriptsize  \textbf{79.07\%}} &  \cellcolor{gray!50}  {\scriptsize  \textbf{86.82\%}}\\
{\scriptsize   Enhanced-SPMF DenseNet-40 (\textbf{ours})}  &   {\scriptsize \textcolor{blue}{\textbf{79.95\%}} } &  {\scriptsize  \textcolor{blue}{\textbf{87.52\%}}}\\
\hline
\end{tabular}
\end{table} 
For the NTU RGB+D dataset, as shown in  Table~\ref{NTU-Accuracy}, the proposed DenseNet-40 achieves an accuracy of 79.95\% on the Cross-Subject and 87.52\% on Cross-View evaluations, respectively. These results demonstrate the effectiveness of the proposed representation and deep learning framework since they surpassed previous state-of-the-art approaches reported in \cite{Vemulapalli2014HumanAR,Du2015HierarchicalRN,Shahroudy2016NTURA,Liu2016SpatioTemporalLW,Zhang2017OnGF,Liu2017EnhancedSV,liu2018skeleton,Wang2017ModelingTD,Hu2015JointlyLH} as well as a higher level of performance than SPMF \cite{Hieu_et_al_2018}. Fig.~\ref{fig:5} (\textit{right}) shows the training loss and test accuracy of the proposed DenseNet-40 on the NTU RGB+D dataset. On the CEMEST dataset, an accuracy of 91.18\% has been made by the DenseNet-40 in the first setting. In the second setting, transfer learning is used. The experimental results show that the proposed method reached an accuracy of 95.70\%, increasing the performance by nearly 5\% compared to the first experiment. This could be explained by the fact that since the CEMEST dataset is quite small, it benefits from the knowledge transfer coming from larger datasets such as the MSR Action3D and NTU RGB+D datasets. This result indicates that the use of data augmentation and transfer learning is crucial to address the small amount of samples in real-world datasets. Fig.~\ref{fig:cemest} shows learning curves of the proposed deep learning networks on the CEMEST dataset from scratch (Fig.~\ref{fig:cemest}a -- Fig.~\ref{fig:cemest}c), pre-training on the combined dataset (Fig.~\ref{fig:cemest}d -- Fig.~\ref{fig:cemest}f) and fine-tuning on CEMEST dataset (Fig.~\ref{fig:cemest}g -- Fig.~\ref{fig:cemest}i).
\subsection{An ablation study on Enhanced-SPMF}\label{oblation-study}
We believe that the use of the smoothing filter and  the AHE algorithm \cite{pizer1987adaptive} helps the proposed representation to be more discriminative, which improves recognition accuracy. To verify this hypothesis, we carried out an ablation study on the proposed representation by removing the color enhancement module and seeing how that affects performance. We observed that this kind of transformation is needed for improving learning performance of deep neural networks. Specifically, we trained the proposed DenseNet-40 on both the SPMFs and Enhanced-SPMFs provided by MSR Action3D dataset \cite{Li2010ActionRB}. During training, the same hyper-parameters and training methodology were applied. The experimental results indicate that the proposed deep network achieves better recognition accuracy when trained on the Enhanced-SPMF (+1.42\%). This result validates our hypothesis above. \newpage
\subsection{Computational efficiency evaluation \\[-0.3cm]}\label{computational-efficiency}
\noindent The proposed learning framework comprises three main stages: (\textbf{1}) the computation of Enhanced-SPMF; (\textbf{2}) the training stage; and (\textbf{3}) the inference stage. To evaluate the computational efficiency of this method, we measure the execution time of each stage on the AS1 subset/MSR Action3D dataset with the proposed DenseNet-40 network, which only has 1.0M parameters. With the implementation in Python using Keras and training on a single GTX Ti 1080 GPU, the training process takes less than one hour to reach convergence. While the inference stage, including the stage (\textbf{1}) that is executed on a CPU and the stage (\textbf{3}), takes an average of 0.175s per sequence without parallel processing. This result verifies the appropriateness of our method in terms of computational cost. Additionally, the computation of the Enhanced-SPMF can be implemented on a GPU for real-time applications.\\[-0.5cm]
\newpage
\section{Conclusions \\[-0.5cm]} \label{sect:Conclusions}
\noindent We introduced a deep learning framework for 3D action recognition from skeletal data. A new motion representation that captures the spatio-temporal patterns of skeleton movements and encodes them into color images has been proposed. Densely connected networks have been designed to learn and recognize actions from the proposed representation in an end-to-end manner. Experiments on two public datasets have demonstrated the effectiveness of our method, both in terms of accuracy as well as computational time. We also introduced CEMEST, a new real-wold surveillance dataset containing both normal and anomalous events for studying human behaviors in public transport. Experimental results on this dataset show that the proposed deep learning based approach achieved promising results. We are currently expanding this study by adding more visual evidence to the network in order to further gains in performance. A new approach for 3D pose estimation will also be studying to replace depth sensors. The preliminary results are encouraging. 
\subsection*{Acknowledgements}
This research was supported by the Cerema, France. Sergio A. Velastin is grateful for funding from the Universidad Carlos III de Madrid, the EU's 7th Framework Programme for Research, Technological Development and demonstration (grant 600371), Ministerio de Economia, Industria y Competitividad (COFUND2013-51509), Ministerio de Educaci\'on, cultura y Deporte (CEI-15-17) and Banco Santander.
\bibliographystyle{splncs04}
\bibliography{refs}
\end{document}